\def\year{2020}\relax
\documentclass[letterpaper]{article} 
\usepackage{aaai20}  
\usepackage{times}  
\usepackage{helvet} 
\usepackage{courier}  
\usepackage[hyphens]{url}  
\usepackage{graphicx} 
\usepackage{graphicx}  
\usepackage{url}            
\usepackage{graphicx} 
\usepackage{subfigure}

\usepackage{algorithm}
\usepackage{algorithmic}

\usepackage{bbm}
\usepackage{bm}
\usepackage{amsmath}
\usepackage{amssymb}
\usepackage{amsthm}
\usepackage{mathrsfs}
\usepackage{graphicx}
\usepackage{algorithm}
\usepackage{algorithmic}
\usepackage{subfigure}
\usepackage{epsfig}
\usepackage{epsf}
\usepackage{booktabs}
\usepackage{tabularx}
\usepackage{tabulary}
\usepackage{multirow}
\usepackage{graphics}
\usepackage{longtable}
\usepackage{subfigure}

\def\e{{\bf e}}

\def\w{{\bf w}}
\def\x{{\bf x}}
\def\y{{\bf y}}

\def\algo{{CWSL}}

\title{Weakly Supervised Learning Meets Ride-Sharing User Experience Enhancement}
\author{Lan-Zhe Guo\textsuperscript{\rm 1}
		Feng Kuang \textsuperscript{\rm 2}
		Zhang-Xun Liu \textsuperscript{\rm 2}
		Yu-Feng Li \textsuperscript{\rm 1}
		Nan Ma \textsuperscript{\rm 2}
		Xiao-Hu Qie \textsuperscript{\rm 2}\thanks{This work was supported by the National Natural Science Foundation of China (61772262) and Didi research fund. Yu-Feng Li is the corresponding author of this work.}\\
		\textsuperscript{\rm 1}{National Key Laboratory for Novel Software Technology, Nanjing University, Nanjing 210023, China}\\
		\textsuperscript{\rm 2}{Didi Chuxing, Beijing, China}\\
		guolz@lamda.nju.edu.cn, \{kuangfeng, liuzhangxun\}@didichuxing.com, \\ liyf@lamda.nju.edu.cn, \{mandyma.edu, tiger.qie\}@didiglobal.com
}
\begin{document}
	
	\maketitle
	
\begin{abstract}
Weakly supervised learning aims at coping with scarce labeled data. Previous weakly supervised studies typically assume that there is only one kind of weak supervision in data. In many applications, however, raw data usually contains more than one kind of weak supervision at the same time. For example, in user experience enhancement from Didi, one of the largest online ride-sharing platforms, the ride comment data contains severe label noise (due to the subjective factors of passengers) and severe label distribution bias (due to the sampling bias). We call such a problem as \emph{`compound weakly supervised learning'}. In this paper, we propose the \algo\ method to address this problem based on Didi ride-sharing comment data. Specifically, an instance reweighting strategy is employed to cope with severe label noise in comment data, where the weights for harmful noisy instances are small. Robust criteria like AUC rather than accuracy and the validation performance are optimized for the correction of biased data label. Alternating optimization and stochastic gradient methods accelerate the optimization on large-scale data. Experiments on Didi ride-sharing comment data clearly validate the effectiveness. We hope this work may shed some light on applying weakly supervised learning to complex real situations.
\end{abstract}

\section{Introduction}
	
Conventional supervised learning has achieved great success in various applications. Most successful techniques, such as deep learning, require ground-truth labels to be given for a big training data set. In many real-world applications, however, it can be difficult to attain strong supervision due to the fact that the hand-labeled data sets are time-consuming and expensive to collect. Thus, it is desirable for machine learning techniques to be able to work well with weakly supervised data~\cite{zhou2017brief}. 
	
Compared to the data in traditional supervised learning, weakly supervised data does not have a large amount of precise label information. Typically, three types of weakly supervised data exist: incomplete supervised data~\cite{oliver2018realistic}, i.e., only a small subset of training data is given with labels whereas the other data remains unlabeled; inaccurate supervised data~\cite{frenay2014classification}, i.e., the given labels have not always been ground-truth and inexact supervised data~\cite{carbonneau2018multiple}, i.e., only coarse-grained labels are given. Weakly supervised learning (WSL) has attracted considerable attention, which has consequently resulted in a large number of WSL methods, e.g., ~\cite{liu2014robust,oliver2018realistic,hendrycks2018using}. 
	
In previous studies on weakly supervised learning, a basic assumption is that there is only one kind of weak supervision in training data. However, in many real-world applications such as ride-sharing user experience enhancement, such an assumption is difficult to hold. In Didi ride-sharing comment data, it suffers from severe label noises, i.e., collected labels from comment questions in ride-sharing user experience are low-quality since they can easily be influenced by subjective factors (e.g., emotions) and many misoperations of the passengers. Meanwhile, there is also severe label distribution bias, i.e., label distributions between training and testing data are different, because the negative comments in ride-sharing user experience are rare and they are dynamically changing with time.
	
It is evident that single type weakly supervised learning could not tackle the problem concerned in this paper. For example, label noise learning methods ignore the label distribution bias which may seriously hurt the performance in practice; while label distribution bias correction methods assume that the ground-truth label is accessible to each instance, which is not the case in our situations. Note that the data scenario studied in this paper is quite different from the previous weakly labeled studies. We call such a problem as \emph{`compound weakly supervised learning'}.
	
In this paper, a novel method named \algo\ is proposed and verified on Didi ride-sharing comment data, where the goal is to pre-detect negative comments and help improve ride-sharing user experience. It is crucial to decrease the influence of noisy instances in the training stage at the same time correct the label distribution bias to make the model work well in practice. In our method, an instance reweight strategy is employed to cope with severe labeling noise by assigning a small weight to noisy instances. To make the learned model performs well on the test distribution, the instance reweight process is under the guidance of validation performance which is evaluated by a robust AUC criteria. The idea is formulated as a novel bi-level optimization. An alternating optimization and stochastic gradient method is adopted to make the method adapt to large-scale data. Experiments on Didi ride-sharing comment data clearly validate the effectiveness. 
	
	
\section{Related Work}
	
The problem focused in this paper, i.e. compound weakly supervised learning, more specifically, is the intersection of label noise learning and label distribution bias problem. Both can be regarded as weakly supervised frameworks where label information conveyed by training instances is incorrect or biased with true distribution.
	
The performance of machine learning models has been shown to degrade noticeably in the presence of label noise~\cite{frenay2014classification}. Considerable efforts have been made to correct or learn with noisy labels~\cite{natarajan2013learning,biggio2011support,frenay2014classification,li2019towards}.  \cite{mnih2012learning} allow for label noise robustness by modifying the model’s architecture. \cite{DBLP:conf/uai/NorthcuttWC17} propose to reweight the noisy instances according to the predicted probability. \cite{patrini2017making} make use of the forward loss correction mechanism, and propose an estimate of the label corruption estimation matrix. There are also methods that correct labels with small clean validation data. For example, \cite{veit2017learning} use validation data to train a label cleaning network by estimating the residuals between the noisy and clean labels. \cite{hendrycks2018using} use validation data to estimate a label correction matrix. However, these methods are not sufficient to conquer the studied problem well, because they ignore the label distribution bias that may cause performance degradation seriously in practice.
	
One kind of strategy to correct label distribution bias is to develop bias robust models. \cite{liu2014robust} develops a framework that can learn a robust bias-aware probabilistic classifier using a minimax estimation formulation. \cite{DBLP:conf/aistats/ChenMLZ16} propose a label distribution bias robust regression method by minimizing conditional Kullback-Leibler divergence and considering the worst-case performance. Another kind of strategy attempts to correct label distribution bias by reweighting training instances. \cite{huang2007correcting} proposes to reweight instances by matching distributions between training and testing sets in feature space. \cite{zhang2013domain} proposes to estimate the weights by reweighting data to reproduce the covariate distribution on the test distribution. There are also methods aim to correct label bias via domain adaptation methods~\cite{Azizzadenesheli19}. However, these methods assume that the ground-truth label is accessible to each instance, which is not the case in our situations and can influence the model performance severely. 
	
Our method is also related to other `mixed' cases under weakly supervised learning frameworks such as \emph{multi-instance multi-label learning}~\cite{zhou2012multi}, \emph{semi-supervised label noise learning}~\cite{DBLP:conf/kdd/ZhangZJZ19}, \emph{semi-supervised multi-label learning}~\cite{wei2018learning}, \emph{semi-supervised weak label learning}~\cite{dong2018learning} and \emph{semi-supervised partial label learning}~\cite{DBLP:conf/ijcai/WangLZ19}. It is worth noting that our paper is different from these works as we focus on the mix of severe label noise and biased label distribution.

\begin{figure*}[h] 
		\centering
		\small
		\includegraphics[scale=0.6]{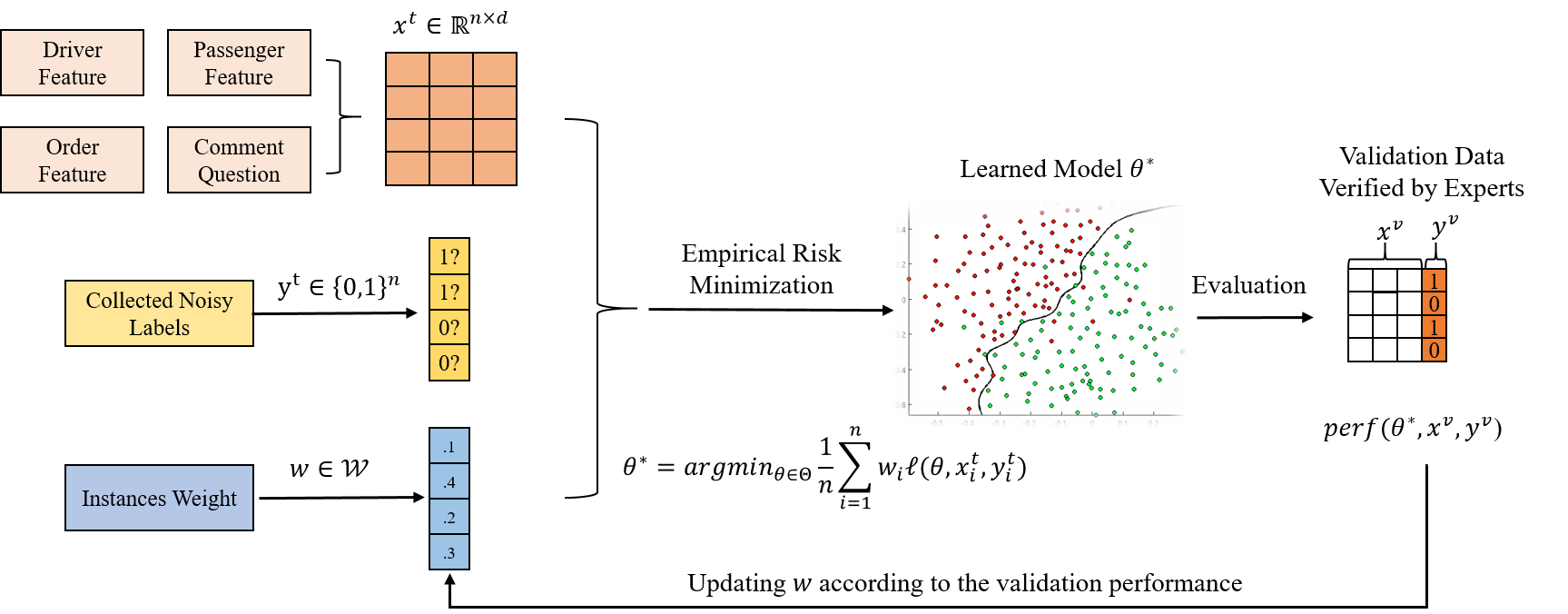}
		\caption{Illustration of the proposed algorithm. }\label{fig:framework}
\end{figure*}
	
	\section{The Proposed Method}
	
	\subsection{Problem Setting and Notations}
	\begin{table}[]
		\centering
		\begin{tabular}{ll}
			\toprule
			Notation  & Meaning \\
			\hline
			$\x \in \mathbb{R}^{d}$ & Feature vector of instance\\
			$y \in \{0, 1\}$ & Label of instance\\
			$n$ & Number of training instances\\
			$m$ & Number of validation instances, $m \ll n$\\
			$\{\x^t_i, y^t_i\}_{i=1}^{n}$ & Training instances\\
			$\{\x^v_i, y^v_i\}_{i=1}^{m}$ & Validation instances\\
			$\theta$ & Model parameters\\
			$\hat{y}=f(\x, \theta)$ & Prediction of model $\theta$ on instance $\x$\\
			$\w\in\mathcal{W}$ & Weights for training instances\\
			$w_i$ & Weight for the $i$-th training instance\\
			$\ell(\cdot, \cdot)$ & Training loss function\\
			$\mathcal{L}(\cdot, \cdot)$ & Validation loss function\\
			$\lambda_{\theta}$ & Step size for $\theta$\\
			$\lambda_{\w}$ & Step size for $\w$\\
			\bottomrule
		\end{tabular}
		\caption{Summary of important notations used in this paper.}
		
		\label{tbl:notations}
	\end{table}
	
Let $(\x, y) \in \mathcal{X} \times \mathcal{Y}$  be the feature-label pair where $\x \in \mathbb{R}^{d}$ is the feature vector including driver features, passenger features, order specific features and comment question, $\y\in\{0,1\}^n$ is the label where 1 indicates positive comment and 0 indicates negative comment. $\{(\x^t_i, y^t_i), 1 \leq i \leq n\}$ is the training set and $\{(\x_i^v,y_i^v), 1 \leq i \leq m\}$ is a small clean unbiased validation set verified by domain experts. In reality, $m \ll n$. Hereafter, we will use superscript $t$ to denote the training set and superscript $v$ to denote validation set. 
	
Let $\hat{y} = f(\theta, \x)$ be the prediction of model $\theta$ on instance $\x$, $\ell(\cdot,\cdot)$ be a loss function to be minimized during the training procedure. Following the classical ERM (Empirical Risk Minimization) framework, we aim to minimize the expected loss for the training set: $\frac{1}{n}\sum_{i=1}^{n}\ell(f(\theta, \x^t_i),y^t_i)=\frac{1}{n}\sum_{i=1}^{n}\ell_i(\theta)$ where each training instance is weighted equally, and $\ell_i(\theta)$ stands for the loss function associating with instances $(\x^t_i, y^t_i)$. The important notations and meanings used in this paper are summarized in Table~\ref{tbl:notations}.
	
\subsection{Instance Reweight Mechanism}
The difficulty of \algo\ is to learn with severe noisy labels and at the same time making the learned model works well on real distribution. One intuitive solution to handle noisy instances is to assign a small weight to noisy instances to decrease their influence~\cite{DBLP:conf/uai/NorthcuttWC17}, i.e., a weighted loss is minimized during the training procedure:
\begin{equation}
	\theta^* = \arg\min_{\theta \in \Theta}\frac{1}{n}\sum_{i=1}^{n}w_i\ell_i(\theta)
\end{equation}
where $w_i$ is the weight for each instance, $\Theta$ is the space of model parameter, $\theta^*$ is the optimal model learned by the ERM procedure.
	
However, $w_i$ is unknown upon beginning. Previous label noise learning studies typically calculate the instances weights based on the training loss for the reason that when the noisy instances are fewer than clean instances, the clean instance is more likely to have a smaller training loss~\cite{frenay2014classification}. In our method, to make the learned model performs well on test distribution, a small clean unbiased validation data is introduced. It is usually acceptable to collect small validation data with small expense in real applications. For example, in Didi, an annotation team can collect accurate labels for a small dataset via telephone interview.
	
We argue that the optimal weight is that the model trained with this weight should maximize the validation performance since the validation performance acts as a good approximation of the generalization performance. Specifically, once the model trained with $\w$ owns a good validation performance, it can possibly perform well on the general data distribution. Formally, we consider the following optimization,
\begin{equation}
\w^* = \arg \max_{\w}perf(\theta^*, \x^v, \y^v)
\end{equation}
where $perf(\theta^*, \x^v, \y^v)$ is the task specific performance measure (e.g., AUC, Accuracy, F1-Score) of model $\theta^*$ on validation set $(\x^v, \y^v)$.
	
Therefore, by combining the above considerations, the objective of our weight learning algorithm is formulated as:
\begin{eqnarray}\label{eq:framework}
&\max\limits_{\w \in \mathcal{W}}& perf(\theta^*, \x^v, \y^v)\\ 
&\mbox{s.t.}&    \theta^* = \arg\min_{\theta \in \Theta}\frac{1}{n}\sum_{i=1}^{n}w_i\ell_i(\theta)\nonumber
\end{eqnarray}
where the weight $\w$ is from a convex set $\mathcal{W}$ that is typically set as $\mathcal{W} = \{\w|0 \leq \w \leq 1\}$. Figure~\ref{fig:framework} illustrates the framework of our method. 
	
The proposed formulation Eq.(\ref{eq:framework}) is a bi-level optimization problem where one optimization problem is nested within another problem~\cite{bard2013practical}. The lower-level optimization is to find a weighted empirical risk minimizer model given the training set whereas the upper-level optimization is to maximize the validation performance given the learned model. The effectiveness of bi-level optimization~\cite{bard2013practical} has been recently demonstrated in the machine learning community. For example, \cite{franceschi2017forward,franceschi2018bilevel} propose hyper-parameter optimization methods based on bi-level formulation. \cite{zhang2018training} introduce validation data and debug the training set labels according to validation performance. \cite{ren2018learning} propose to reweight examples for robust deep learning based on bi-level optimization meta-learning. 
	
\subsection{Robust AUC Criteria}
	
To conquer the bias label distribution, where the label distributions in the training data and testing data are different since negative comments in ride-sharing user experience are much fewer than that of positive comments while they are dynamically changing with time in test data. This means directly training a binary classifier in terms of accuracy will predict almost all incoming instances to be positive ones, which obviously does not make sense. So we adopt AUC as the performance measure as AUC is not sensitive to the class ratio~\cite{liu2008exploratory}, which has been shown effective for many applications, such as website ad click prediction~\cite{mcmahan2013ad} where only a very small fraction of web history contains ads clicked by visitors.
	
Specifically, let $\x^+, \x^-$ be positive and negative instances generated from distribution $\mathcal{P}^+$ and $\mathcal{P}^-$. The AUC is defined as:
\begin{equation}\label{eq:auc}
\text{AUC} = \mathbb{E}_{\begin{subarray}{c} \x^+ \sim \mathcal{P}^+ \\ \x^- \sim \mathcal{P}^-\end{subarray}}[\mathbb{I}\{f(\theta, \x^+) - f(\theta, \x^-) > 0\}]
\end{equation}
where $\mathbb{I(\cdot)}$ is the indicator function which returns 1 if the argument is true and 0 otherwise. This expectation is the probability that a positive instance is ranked higher than a negative instance.
	
Owing to its non-convexity and discontinuous, direct optimization of AUC often leads to a NP-hard problem~\cite{gao2013one}. To make a compromise for avoiding computational difficulties, we propose to optimize a convex surrogate loss function~\cite{DBLP:journals/corr/abs-1805-11221}. 
	
Replacing $\mathbb{I}\{f(\theta, \x^+) - f(\theta, \x^-) > 0\}$ in Eq.(\ref{eq:auc}) with the surrogate loss $\phi(f(\theta,\x^+) - f(\theta,\x^-))$, we now aim to minimize:
\begin{equation}\label{eq:phi-risk}
\mathbb{E}_{\begin{subarray}{c} \x^+ \sim \mathcal{P}^+ \\ \x^- \sim \mathcal{P}^-\end{subarray}}[\phi(f(\theta, \x^+) - f(\theta, \x^-))]
\end{equation}

In this paper, we adopt the pairwise squared loss $\phi(t) = (1-t)^2$ as the surrogate loss function, for the reason that it is convex differentiable and we can easily prove its consistency with the original AUC according to Theorem 2 in~\cite{DBLP:conf/ijcai/GaoZ15}.
	
Denote the positive and negative instances in validation set $\x^v$ as $\mathcal{S}^+=\{\x_1^{v+}, \cdots, \x_{m_+}^{v+}\}$ and $\mathcal{S}^-=\{\x_1^{v-}, \cdots, \x_{m_-}^{v-}\}$ where $m_+$ and $m_-$ are number of positive and negative instances. For a model $\theta$, the surrogate validation loss can be written as:
	\begin{align}\label{eq:upper-level}
	\nonumber
	\mathcal{L}(\theta)&=\frac{1}{m^+m^-}\sum_{i=1}^{m^+}\sum_{j=1}^{m^-}[(1-(f(\theta,\x^{v+}_i) - f(\theta,\x^{v-}_j)))^2] \\ \nonumber
	&=\frac{1}{m^+m^-}\sum_{i=1}^{m^+}\sum_{j=1}^{m^-}[(1-(\theta^{\top}\x^{v+}_i-\theta^{\top}\x^{v-}_j))^2]\\ \nonumber
	&=1-2\theta^{\top}[\frac{1}{m^+m^-}\sum_{i=1}^{m^+}\sum_{j=1}^{m^-}(\x_i^{v+}-\x_j^{v-})] \\ \nonumber 
	&+\theta^{\top}[\frac{1}{m^+m^-}\sum_{i=1}^{m^+}\sum_{j=1}^{m^-}(\x_i^{v+}-\x_j^{v-})(\x_i^{v+}-\x_j^{v-})^{\top}]\theta\\ 
	&=1-2\theta^{\top}\mu_m + \theta^{\top}\Sigma_m\theta
	\end{align}
where 
	\small{
		\begin{eqnarray}
		&\mu_m &= \frac{1}{m^+m^-}\sum_{i=1}^{m^+}\sum_{j=1}^{m^-}(\x^{v+}_i - \x^{v-}_j)\\
		&\Sigma_m &= \frac{1}{m^+m^-}\sum_{i=1}^{m^+}\sum_{j=1}^{m^-}(\x^{v+}_i-\x^{v-}_j)(\x^{v+}_i-\x^{v-}_j)^{\top} 
		\end{eqnarray}}
	
	\subsection{Gradient Descent Updating}
	
	By combining the forms Eq.(\ref{eq:framework}) and Eq.(\ref{eq:upper-level}), the final objective of the optimization is formulated as:
	\small
	\begin{eqnarray}\label{eq:obj}
	&\min\limits_{\w \in \Omega}& \mathcal{L}(\theta^*)\\
	&\mbox{s.t.}&      \theta^* = \arg\min\limits_{\theta \in \Theta} \frac{1}{n}\sum_{i=1}^{n}w_i\ell_i(\theta)\nonumber
	\end{eqnarray}
	where $\mathcal{L}(\theta^*)=-2\theta^{*\top}\mu_m + \theta^{*\top}\Sigma_m\theta^*$ and the negative log-likelihood is adopted as the loss function in our task: $\ell_i(\theta) = -y^t_i\theta^{\top}\x^t_i + \log(1+\e^{\theta^{\top}\x^t_i})$.
	
	Notice that the lower-level problem can be replaced equivalently with its Karush-Kuhn-Tucker (KKT) condition~\cite{boyd2004convex}:
	\begin{equation}\label{eq:kkt}
	g(\w, \theta) \equiv \frac{1}{n}\sum_{i=1}^{n}w_i\frac{\partial \ell_i(\theta)}{\partial \theta}= 0
	\end{equation}
	
	According to the implicit function theorem~\cite{DBLP:journals/fm/NakashoFS17}, we have the following Jacobian matrix:
	\begin{equation} \label{eq:J}
	J = \frac{\partial \theta}{\partial \w} =   -\left[
	\begin{matrix}
	\frac{\partial g_1}{\partial \theta_1} & \cdots & \frac{\partial g_1}{\partial\theta_d} \\
	\vdots &  & \vdots \\
	\frac{\partial g_d}{\partial\theta_1} & \cdots & \frac{\partial g_d}{\partial\theta_d}
	\end{matrix} \right]^{-1}\left[
	\begin{matrix}
	\frac{\partial g_1}{\partial w_1} & \cdots & \frac{\partial g_1}{\partial w_n} \\
	\vdots &  & \vdots \\
	\frac{\partial g_d}{\partial w_1} & \cdots & \frac{\partial g_d}{\partial w_n} 
	\end{matrix} \right]
	\end{equation}
	
	Referring to Eq.(\ref{eq:kkt}), we can compute $\frac{\partial g}{\partial \theta}$ and $\frac{\partial g}{\partial \w}$ as:
	\begin{equation}\label{eq:theta_and_w}
	\frac{\partial g}{\partial \theta}= \w^{\top}\left(\frac{\partial^2 \ell(\theta)}{\partial \theta \partial \theta^{\top}}\right),\;\;\;\;
	\frac{\partial g}{\partial \w}= \frac{\partial \ell(\theta)}{\partial \theta}
	\end{equation}
	
	The gradient and Hessian of the training loss are posed as:
	\begin{equation}
	\frac{\partial \ell(\theta)}{\partial \theta}= \frac{1}{n}\sum_{i=1}^{n}\x_i^t(-y^t_i + p(\x^t_i;\theta))
	\end{equation}	
	\begin{equation}
	\frac{\partial^2 \ell(\theta)}{\partial \theta \partial \theta^{\top}} =\frac{1}{n}\sum_{i=1}^{n}\x^t_i\x_i^{t\top}p(\x^t_i;\theta)(1-p(\x^t_i;\theta))
	\end{equation}
	where $p(\x^t_i;\theta) = \text{Pr}\{\hat{y}=1|\x^t_i, \theta\}$. 
	
	The matrix $J$ tells us how the model parameters $\theta$ varies with respect to an infinitesimal change to $\w$. Now, we can apply the chain rule to get the gradient of the whole optimization problem w.r.t. $\w$:
	\begin{equation}\label{eq:gradient-w}
	\frac{\partial \mathcal{L}(\theta^*)}{\partial \w} = \frac{\partial \mathcal{L}(\theta^*)}{\partial \theta}\frac{\partial \theta}{\partial \w}
	\end{equation}
	
	Finally, we can solve the bi-level optimization problem with gradient-based methods and the iteration rule is: 
	\begin{equation}
	\w_t = \w_{t-1} - \lambda_{\w}\left.\frac{\partial \mathcal{L}(\theta^*)}{\partial \w}\right|_{\w=\w_{t-1}}
	\end{equation}
	where $\lambda_{\w}$ is the step size for $\w$.
	
	Notice that to solve the gradient w.r.t. $\w$, we need to compute the inverse of the Hessian matrix and in practice, the matrix inverses is not pleasing. Here, we propose an alternative method. We compute $-(\frac{\partial g}{\partial \theta})^{-1}\frac{\partial g}{\partial \w}$ as it is the solution to $(\frac{\partial g}{\partial \theta})\x = -\frac{\partial g}{\partial\w}$. The linear system can be solved conveniently and only requires matrix-vector products, that is we do not have to materialize the Hessian.
	
	\subsection{Efficient Alternating Optimization}
	
	Calculating the optimal $\w$ requires two nested loops of optimization, to further accelerate the optimization for large-scale data, we propose an alternating optimization method by updating $\w$ and $\theta$ iteratively. Instead of updating $\w$ based on the optimal $\theta^*$, we update $\w$ after each iteration of the optimization of $\theta_t$, i.e.,
	\begin{eqnarray}
	&\theta_t &= \theta_{t-1} - \lambda_{\theta}\w_{t-1}\left.\frac{\partial \ell(\theta)}{\partial \theta}\right|_{\theta=\theta_{t-1}} \\
	&\w_t &= \w_{t-1} - \lambda_{\w}\left. \frac{\partial \mathcal{L}(\theta_t)}{\partial \w}\right|_{\w=\w_{t-1}}
	\end{eqnarray}
	Then, compared to regular optimization methods, our method only needs $2\times$ training time. The overall optimization procedure is summarized in Algorithm~\ref{algo:approx}. 
	
	\begin{algorithm}[!t]
		\caption{Alternating Optimization Procedure of \algo.}\label{algo:approx}
		\textbf{Input}: Training set $\{\x^t_i, y_i^t\}_{i=1}^{n}$, validation set  $\{\x^v_i, y^v_i\}_{i=1}^{m}$, current values of weights $\w_0$ and model parameters $\theta_{0}$.
		
		\textbf{Output}: Learned instance weights $\w$ and model parameters $\theta$.
		\begin{algorithmic}[1]
			\FOR {$t = 1$ to $T$}
			\STATE $\theta_t = \theta_{t-1} - \lambda_{\theta}\w_{t-1}\left.\frac{\partial \ell(\theta)}{\partial \theta}\right|_{\theta=\theta_{t-1}} $
			\STATE Compute $\frac{\partial g}{\partial \theta},\;\;\frac{\partial g}{\partial \w}$ with Eq.(\ref{eq:theta_and_w})
			\STATE Compute $J$ with Eq.(\ref{eq:J}).
			\STATE Compute $\frac{\partial \mathcal{L}(\theta_t)}{\partial \w}$ with Eq.(\ref{eq:gradient-w}).
			\STATE $\w_t = \w_{t-1} - \lambda_{\w}\left. \frac{\partial \mathcal{L}(\theta_t)}{\partial \w}\right|_{\w=\w_{t-1}}$.
			\STATE Project $\w_t$ to the constrained set $\mathcal{W}$.
			\ENDFOR
			\RETURN $\w_T$ and $\theta_T$.
		\end{algorithmic}
	\end{algorithm}

	\section{Experiments}  
	In this section, we conduct experiments to demonstrate the effectiveness of the proposed method.
	We first conduct experiments on UCI dataset \emph{breast\_cancer} with synthetic noisy labels to show the instance weights evolution during the training procedure. Then, we present the evaluation settings on real ride comment user experience data in Didi ride-sharing platform include data set preparation, feature transformation, a brief description of the related comparison method and the detail parameter settings. Finally, extensive results of different methods are presented along with the analysis. 
	
	\begin{figure*}
		\subfigure[After 1 Epoch]{
			\begin{minipage}[t]{0.33\linewidth}
				\includegraphics[width=\textwidth]{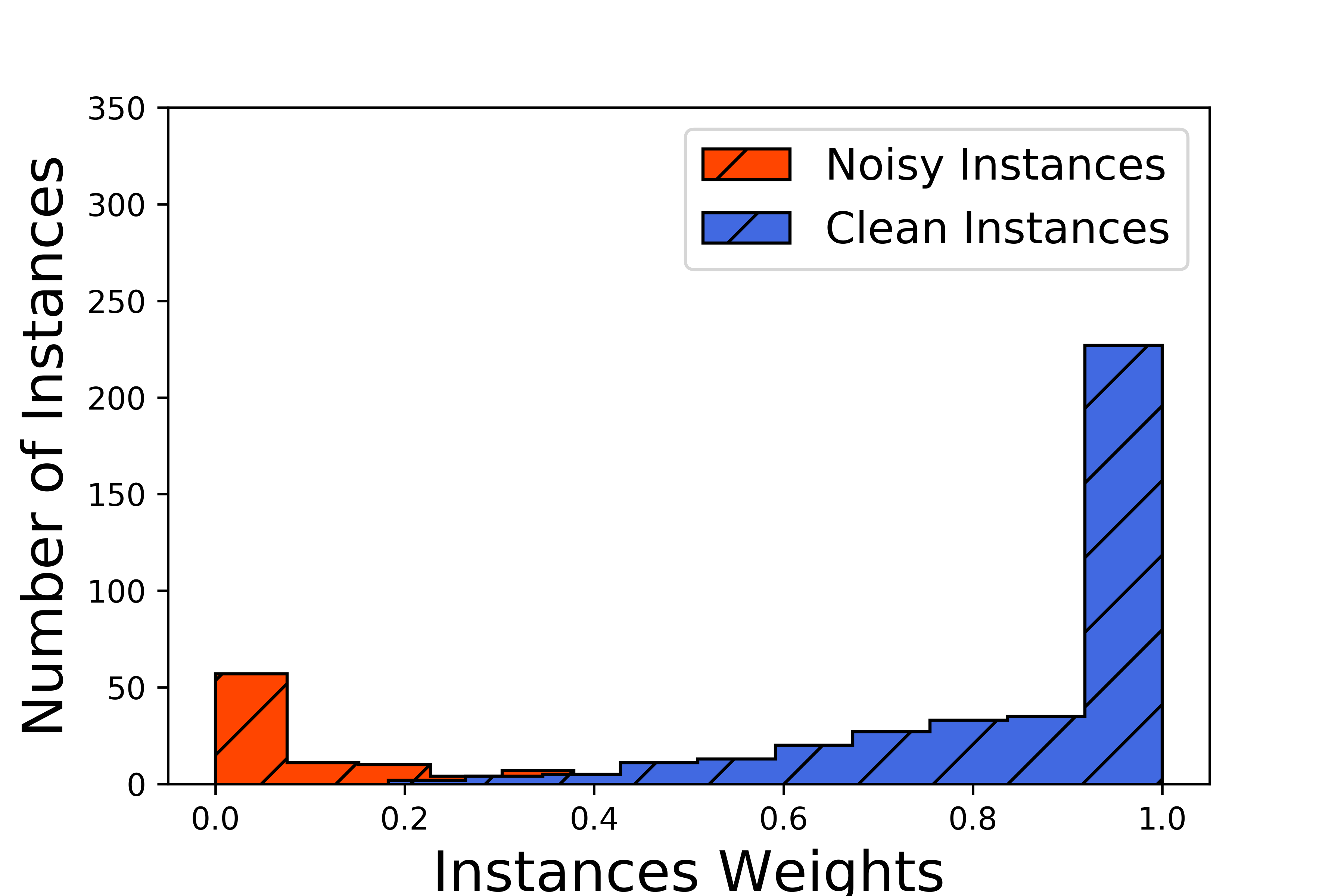}
			\end{minipage}
		}
		\subfigure[After 10 Epochs]{
			\begin{minipage}[t]{0.33\linewidth}
				\includegraphics[width=\textwidth]{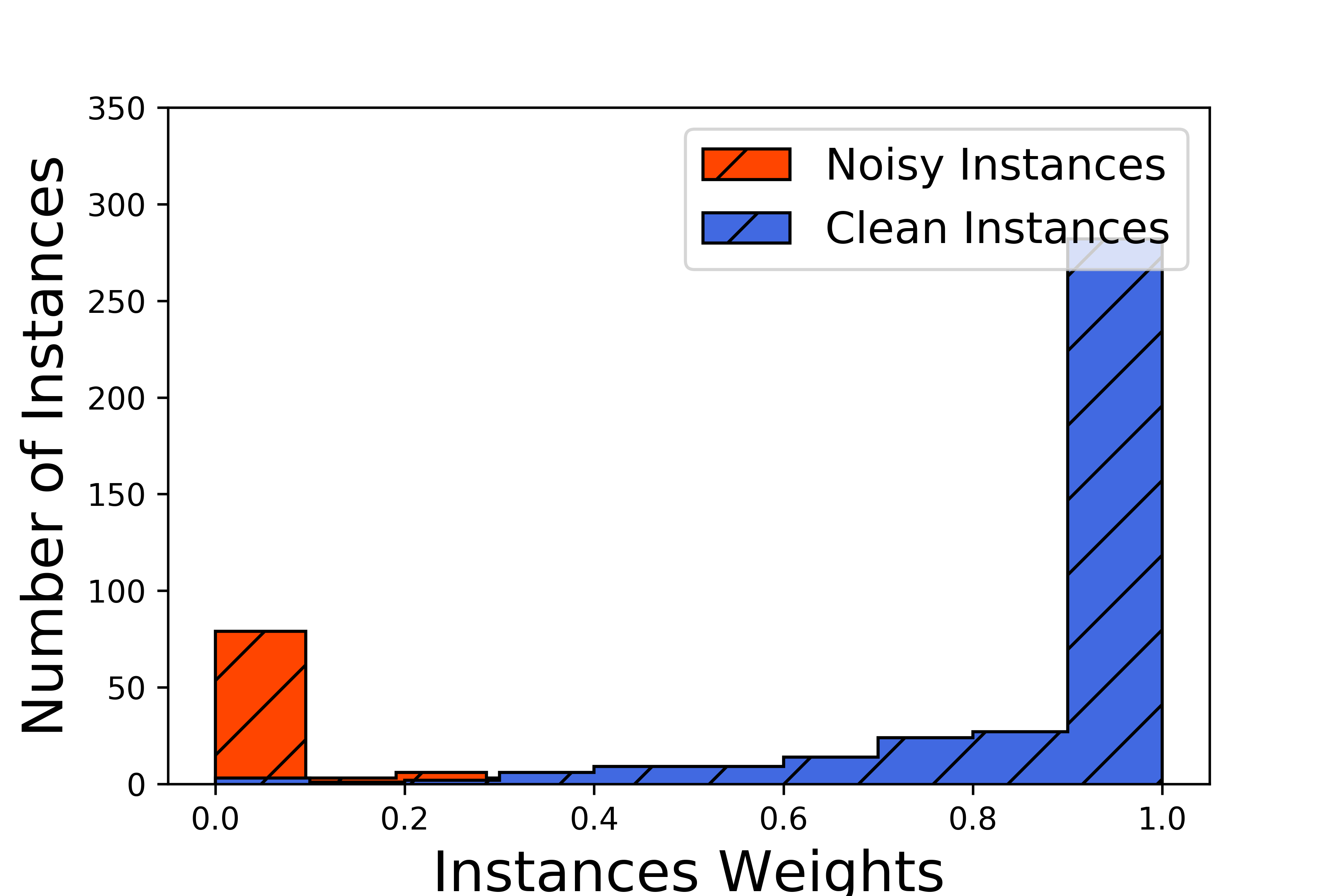}
			\end{minipage}
		}
		\subfigure[After 100 Epochs]{
			\begin{minipage}[t]{0.33\linewidth}
				\includegraphics[width=\textwidth]{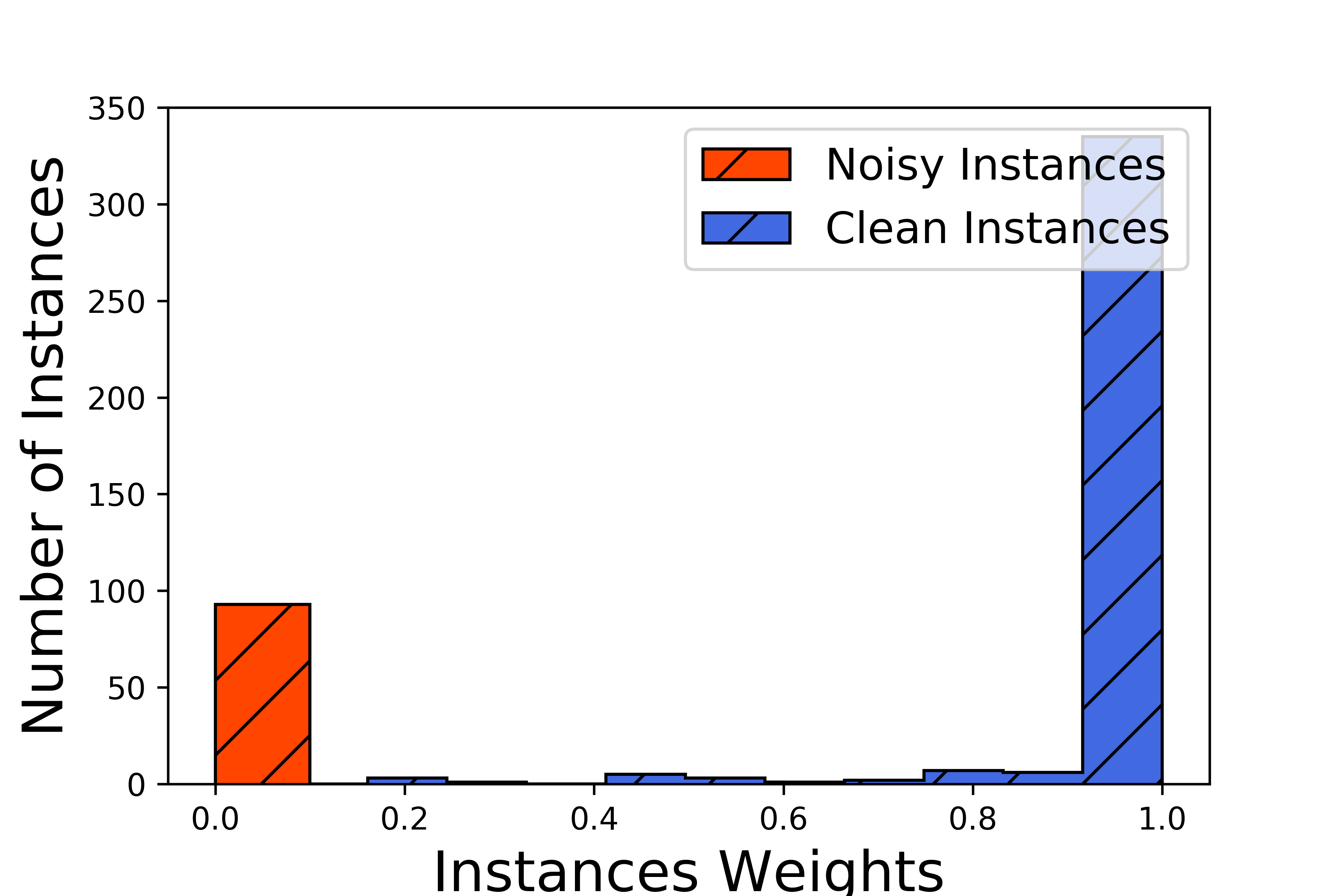}
			\end{minipage}
		}
		\caption{Instances weights distribution after 1 Epoch, 10 Epochs, and 100 Epochs. The instances weights evolution shows that our proposal successfully identifies noisy instances and can reduce their influences by assigning them low weights.}
		\label{fig:synthetic}
	\end{figure*}
	
	\subsection{Synthetic Data}
	To understand how our proposed algorithm contributes to decreasing the influence of noisy labels, we conduct an experiment on UCI dataset \emph{breast\_cancer}~\footnote{https://archive.ics.uci.edu} with synthetic noisy labels to show how the instance weights evolved during the training procedure.
	
	\emph{Breast\_cancer} is a classic binary classification dataset with $569$ instances and $32$ attributes. We split the dataset into 3 parts: 469 training data, 50 validation data, and 50 test data. For the training data, we randomly select 100 instances and reverse their labels as noisy instances.
	
	For this experiment, we adopt a two-layer neural network as the classification model and binary cross-entropy loss to evaluate the validation performance instead of the AUC we described before, as this is no label distribution bias problem. The training epoch is set to 100 and the initial weights of all instances are set to 0.5.
	
	Figure~\ref{fig:synthetic} shows the instance weights evolution during the training procedure. Figure~\ref{fig:synthetic}(a) shows that our method can separate the noisy and clean instances after only 1 epoch. After 10 epochs, the weight of clean instances has been mainly concentrated in 1.0. After 100 epochs, as expected the algorithm nicely separates the two sets and correctly pushes all 100 noisy instances into zero weight while keeping the clean instances into one weight. Finally, the algorithm can achieve $100\%$ accuracy on the test data. These demonstrate that our proposal is able to reliably decrease the influence of noisy instances by assigning them small weights.
	
	\subsection{Real Didi Ride-Sharing Comment Data}
	Didi is one of the largest online ride-sharing platforms that offer peer-to-peer ride-sharing services. Once passengers enter start location and destination, the platform will match a driver nearby to pick up the passenger. The platform provides a comment question to passengers after each order, e.g., ``Was the car smelly?" or ``Did your driver detour?". The goal is to rank these candidate comment questions according to the probability that may receive negative comments because negative comments can help discover the needs of users and the deficiency of services, thus improving service quality and user experience. 
	
	The benchmark ride-sharing comment user experience data set was constructed from the real comments in the main city zone of ride-sharing orders within the time period from Mar 1st, 2019 to Apr 1st, 2019. We sample 600,000 instances as the training data and the training data have severe label noise as we mentioned before. To construct validation and test data, we sample 15,000 data from Apr 2st, 2019 according to the real distribution, and verified their labels by domain experts. Class ratio of negative to positive in the real distribution is approximately 1:30. The verified data is equally split into the validation set and test set. Thus, we have prepared the training data set and clean unbiased validation and test data set.
	
	\subsubsection{Feature Engineering}
	Each instance in the data set is described with hundreds of raw features including driver features, passenger features, order specific features and comment question. For example, driver and passenger features include age, gender, number of negative reviews for a comment question, etc. Order specific features include cities, price, destination, etc. Different comment question is also described with different numerical features.
	
	The raw feature contains both continuous and categorical features, simply train models on the raw feature will leading to poor performance and manual feature engineering is time-consuming. Inspired by~\cite{he2014practical}, we propose to transform the input features of the classifier in order to improve its performance by boosted decision trees. The boosted tree is a powerful and very convenient way to implement non-linear and tuple transformations and has been widely used in Click-Through Rate (CTR) prediction~\cite{richardson2007predicting}.
	
	Specifically, we treat each individual tree as a categorical feature that takes as value the index of the leaf an instance ends up falling in, then we normalize the transformed feature into zero mean and unit variance. The boosted decision tree we
	used follows the XGBoost~\cite{chen2016xgboost}.
	
	\subsubsection{Label Bias Learning Methods}
	To demonstrate the effectiveness of our proposal, we compare the proposed method with three classical classification methods with label distribution bias correction:
	\begin{itemize}
		\item \textbf{XGBoost}~\cite{chen2016xgboost}: which is a scalable machine learning system for tree boosting~\cite{friedman2001greedy}. The effectiveness of XGBoost has been widely recognized in a number of machine learning and data mining challenges, like Kaggle competitions. For XGBoost, we set the weight of positive class into 4 and other parameters are set as default.
		\item \textbf{LR (Logistic Regression)}: which is a baseline method that commonly used in classification tasks. For the Logistic Regression, we adopt the implementation in Sklearn~\footnote{https://scikit-learn.org} and the parameters are set as default.
		\item \textbf{DNN (Deep Neural Network)}: We also design a DNN structure including three hidden layers and a prediction layer, where ReLU~\cite{glorot2011deep} is adopted as the activation function for each hidden layer. The hyper-parameters are tuned based on the validation set. For instance, the number of units for each hidden layer is set to 64. Dropout rate is set as 0.5. We adopt the binary cross-entropy loss and SGD algorithm with learning 0.01 to train the DNN model.
	\end{itemize}
	For these methods, we correct the predicted probability to deal with the label distribution bias problem. According to \textbf{probability calibrating method}~\cite{dal2015calibrating} which is a commonly used method to correct label distribution bias, let $\beta$ be the sampling ratio of the positive instances in training data, $0 < \beta < 1$, the predicted probability of instance $\x$ belong to negative class is $p_s$, the probability can be corrected with $p = \frac{p_s\beta}{p_s\beta-p_s+1}$. 
	
	\subsubsection{Label Noise Learning Methods}
	We also compare our proposal with three state-of-the-art label noise learning methods:
	\begin{itemize}
		\item \textbf{Rank Pruning}~\cite{DBLP:conf/uai/NorthcuttWC17}: Rank pruning is a novel label noise learning method that can estimate the fraction of mislabeling in both the positive and negative training instances and can be applied to any probabilistic classifier. Rank Pruning achieves perfect noise estimation and equivalent expected risk as learning with correct labels in theoretical and achieves nearly the same performance as learning with correct labels. For rank pruning method, we adopt the XGBoost as the basic classification model.
		
		\item \textbf{GLC (Gloden Loss Correction)}~\cite{hendrycks2018using}: Similar to our proposal, GLC also introduces a small set of trusted data and estimate a noise correction matrix based on the validation data. The GLC method achieves state-of-the-art performance on varies real-world dataset, e.g., CIFAR-10 and IMDB. For GLC, we adopt the DNN as the basic model with structure as we described before.
		
		\item \textbf{LTR (Learning to Reweight Examples)}~\cite{ren2018learning}: LTR aims to reweight training instances based on the validation loss. Different from our proposal, LTR simply optimizes the cross-entropy loss on validation data based on a trivial optimization method while our method  optimizes the robust AUC. For LTR method, we adopt the same architecture as the DNN method. However, due to the validation data is class imbalance, simply using LTR will obtain a degenerate solution, i.e., predict the labels of all instances as positive. Therefore, we rescale the validation loss based on ratio 1:30 according to the idea of cost-sensitive learning~\cite{zhou2006multi}.
	\end{itemize}
	
	For methods that fail to utilize validation data (e.g., XGBoost, LR, DNN, Rank Pruning), we add the validation data into the training set to ensure fairness. All methods are trained on the transformed feature. The model structure we adopted is same with DNN and the hyper-parameters of the proposed method are tuned according to the AUC score on the validation set. The initial weights of all instances are set to 0.5, and the training epochs are set to 100. The other two hyper-parameters (step sizes $\lambda_\theta$ and $\lambda_{\w}$), are fixed to $0.1$ and $0.4$, respectively. 
	
\begin{table}[htbp]
	\centering
	\caption{Comparison results of the compared methods. For each method, we report the average AUC score and the running time of 10 runs on the test set.}
	\label{tbl:results}
	\begin{tabular}{c|c|c}
		\toprule
		Methods & AUC & Time Cost (s) \\
		\hline
		
		LR & 78.64 $\pm$ 1.23 & 47.8 \\
		
		DNN & 77.85 $\pm$ 1.17 & 90.4  \\
		
		XGBoost & 79.55 $\pm$ 1.13 & 324.3 \\
		
		Rank Pruning & 82.22 $\pm$ 0.53 & 601.4 \\
		
		GLC & 82.63 $\pm$ 0.49 & 300.5 \\
		
		LTR & 85.45 $\pm$ 0.51 & 211.6 \\
		\hline
		Proposal & \textbf{91.24 $\pm$ 0.47} & 107.1 \\
		
		\bottomrule
	\end{tabular}
\end{table}

	\subsubsection{Comparison Results}
	Table~\ref{tbl:results} shows the experimental results of our proposal and compared methods. From Table~\ref{tbl:results}, we can see that our proposal significantly outperforms all the compared methods. Compared with XGBoost, we achieve more than $12\%$ performance gain and compared with LTR we also achieve nearly $6\%$ performance gain. These show that the previous works that consider only one weak supervision can not deal with the complex compound situations well. Moreover, training models using validation data only cause poor performance (AUC $60.98$ with XGBoost) as the validation data is insufficient to train a good model, so we do not report the result in Table~\ref{tbl:results}. We also try to correct label bias for label noise learning methods but do not achieves performance improvement. One possible reason is that these methods modify the priors of the training set and consequently simply adopt the label bias correction method is not useful. These results demonstrate the effectiveness of our proposal. 

\subsubsection{Validation Set}
	\begin{figure}[htbp] 
		\centering
		\includegraphics[width=3in]{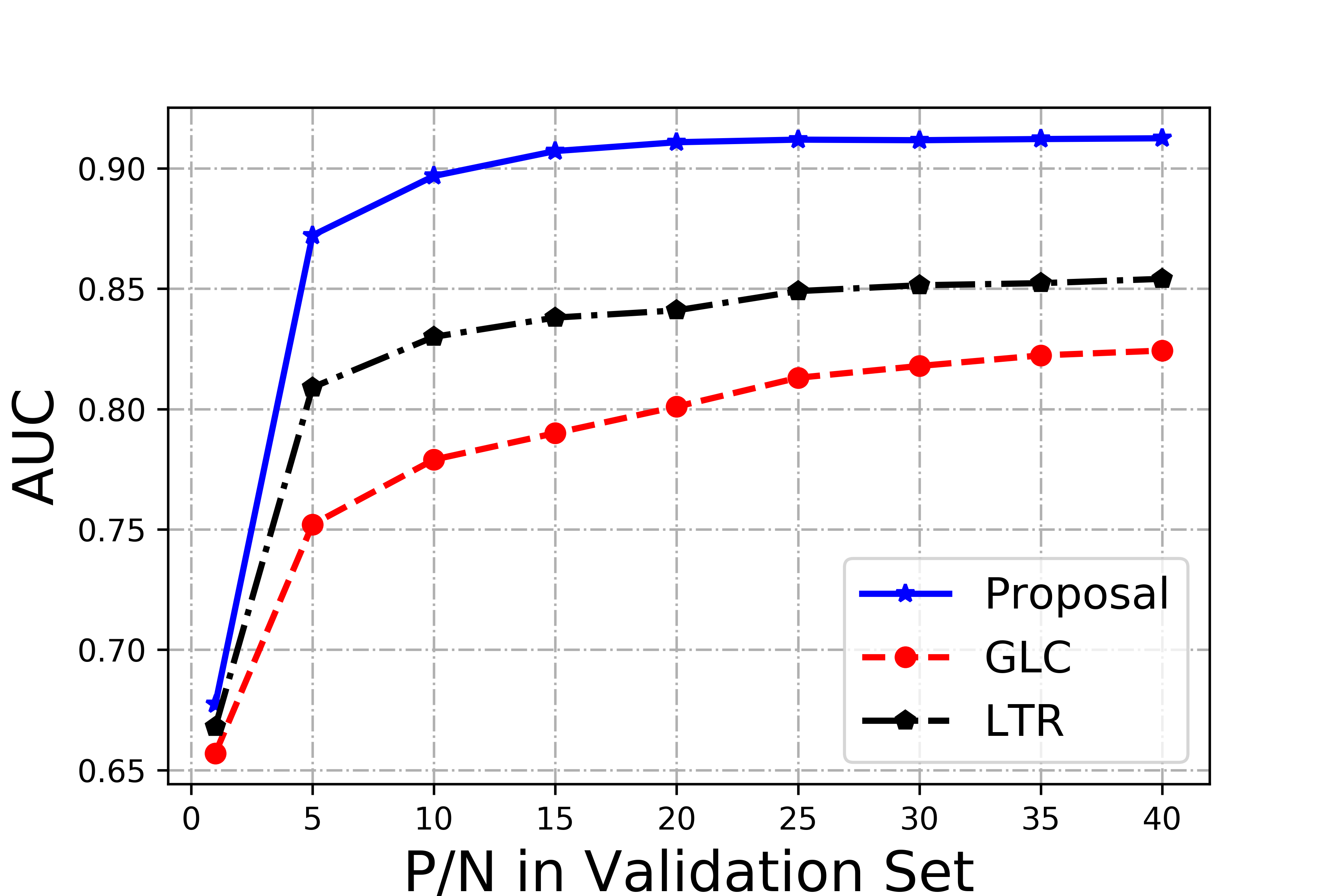}
		\caption{AUC on test data with the varied ratio between positive  to negative class in the validation set}
		\label{fig:ratio}
	\end{figure}
	Previously, experiments are conducted based on clean unbiased validation data. It is interesting to study how the performance influenced by the validation data bias. We conduct experiments with the ratio of positive to negative class in the validation set varied from 1:1 to more than 40:1 on three methods based on the validation set, i.e., LTR, GLC, and our proposed method. The AUC on test data is reported in Figure~\ref{fig:ratio}. From Figure~\ref{fig:ratio}, we can see that, as the change of validation set bias, our proposal is consistently better than the other two well-performed methods. This demonstrates that our proposal is robust to the validation set bias.

\subsubsection{Running Time}
	We also report the running time of compared methods in Table~\ref{tbl:results}. The training of our proposal is implemented in Pytorch~\footnote{\url{https://pytorch.org/}} and all the code are run on a single machine with 10cores 2.20GHz Intel Xeon(R) CPU and 32GB main memory. 
	
	From the results, we can see that our proposal is the fastest method compared with label noise learning methods whereas Rank Pruning that based on XGBoost is the slowest algorithm and our proposal only cost a half time compared with LTR, though these two algorithms are all based on bi-level optimization. Compared with the three baseline methods, our proposal is also comparable with DNN and faster than XGBoost. These demonstrate that our proposal is efficient to handle large-scale dataset.

\setlength{\tabcolsep}{8pt}
\begin{table}[htbp]
	\centering
	\caption{Performance results of our proposal and compared methods on Precision, Recall and F1-Score metrics.}
	\label{tbl:perf-measures}
	\begin{tabular}{c|cc|c}
		\toprule
		Methods & Precision & Recall & F1-Score \\
		\hline
		LR & 24.08\% & 34.69\% & 26.43\% \\
		
		DNN & 23.26\% & 35.34\% & 25.71\%  \\
		
		XGBoost & 9.51\% & 69.80\% & 16.73\% \\
		
		Rank Pruning & 12.26\%  & 54.73\% & 18.27\%\\
		
		GLC & 24.13\% & 34.24\% & 26.79\% \\
		
		LTR & 8.63\% & 87.35\% & 15.70\% \\
		\hline
		Proposal & 16.50\% & 80.82 \% & \textbf{27.41}\%\\
		
		\bottomrule
	\end{tabular}
\end{table}

	\subsubsection{Performance Measures}
	
	It is meaningful to study whether the proposal is effective in other performance measures. Table~\ref{tbl:perf-measures} presents the experimental comparison on Precision, Recall, and F1-Score metrics, that are commonly used measures to evaluate the ranking quality. 
	
	Specifically, denoting all ground-truth positive instances as $G$ and all predicted positive instances as $T$, then precision and recall are defined as follows:
	\begin{equation*}
	precision = \frac{|G\bigcap T|}{|G|},\;\;\;\; recall = \frac{|T\bigcap G|}{|T|}
	\end{equation*}
	and the F1-Score is defined as:
	\begin{equation*}
	F1= \frac{2*precision*recall}{precision+recall}
	\end{equation*}
	
	From the results, we can see that LTR achieve the best recall while the lowest precision, DNN, LR, and GLC achieve similar results on these three metrics. Our proposal trades off precision and recall best and achieves the best result on F1-Score. These results reveal that although the proposal is designed to optimize AUC, it has a certain degree of robustness to the change of performance measures.

	\begin{figure}[htbp]
		\begin{minipage}[t]{0.4\linewidth}
			\includegraphics[width=1.7in]{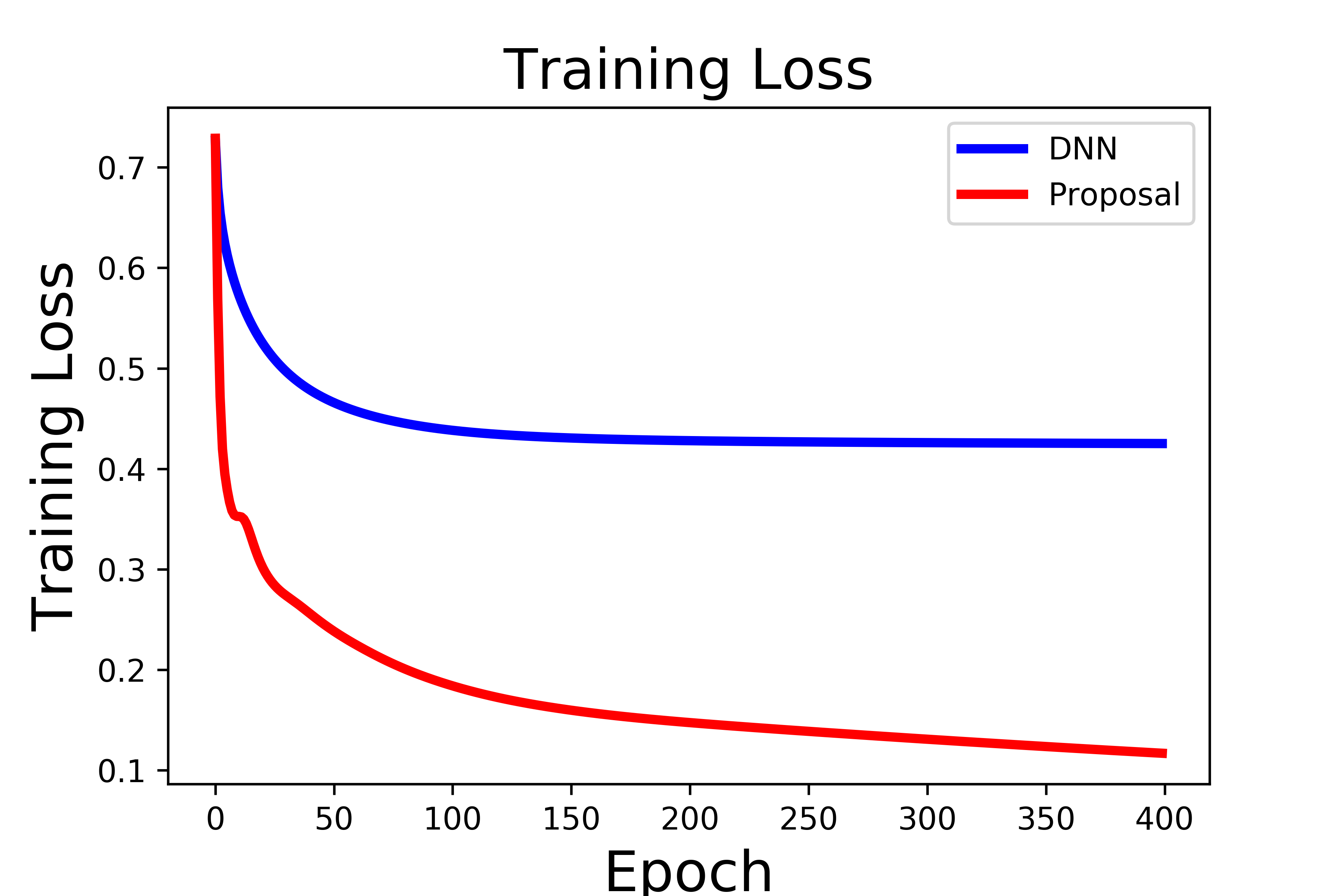}
		\end{minipage}
		\;\;\;\;\;
		\begin{minipage}[t]{0.4\linewidth}
			\includegraphics[width=1.7in]{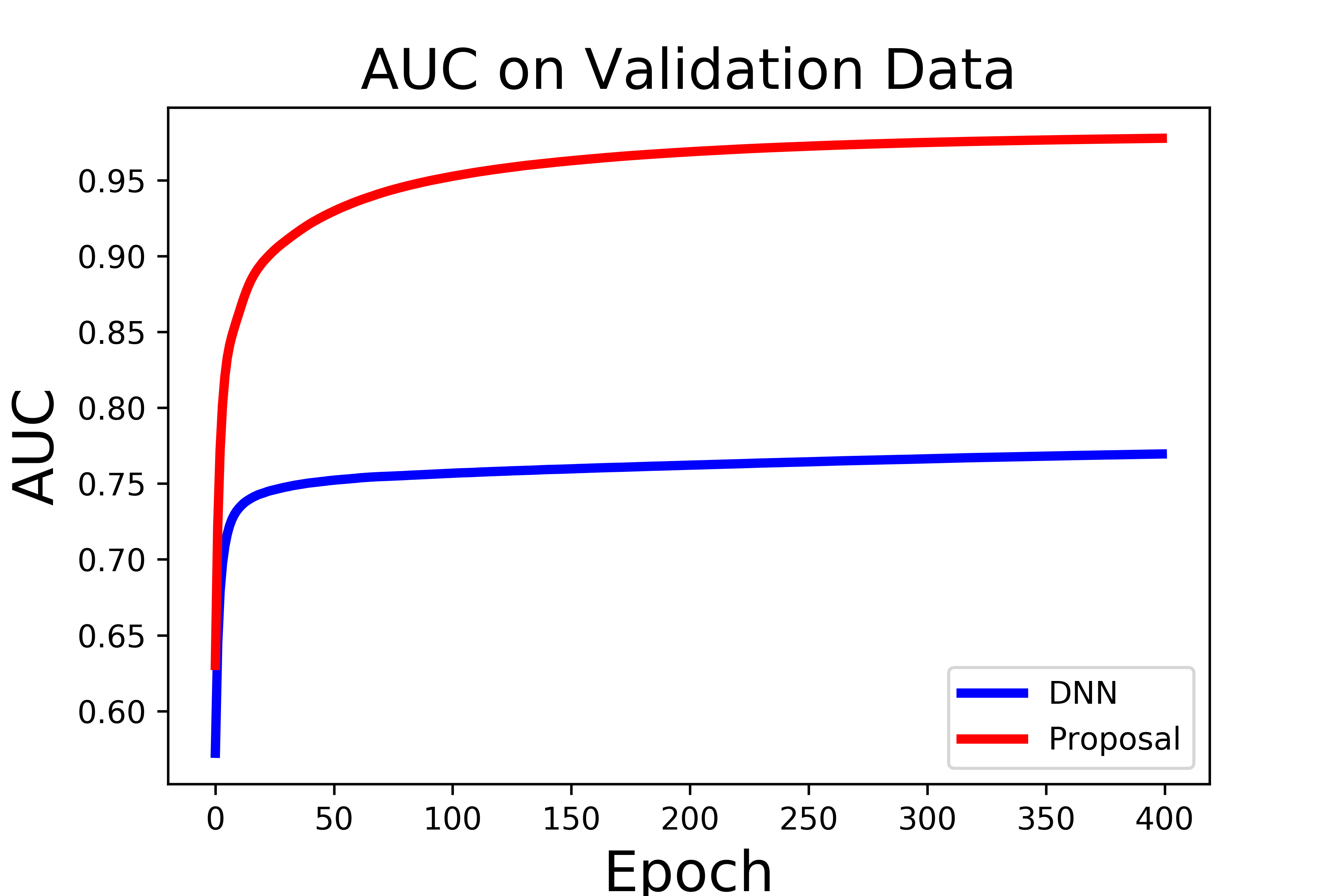}
		\end{minipage}
		\caption{Training Loss v.s. Epoch (left) and Validation AUC v.s. Epoch (right) of our proposal and DNN.}
		\label{fig:convergence}
	\end{figure}
	
	\subsubsection{Convergence}
	
	We also conduct experiments to study the convergence property of our proposed alternating optimization method compared with the normal gradient method. The training loss and validation AUC v.s. training epoch of our proposal and deep neural network are reported in Figure~\ref{fig:convergence}. From Figure~\ref{fig:convergence}, we can see that the number of epoch required to converge of our proposal is comparable with the naive DNN based on single-level gradient descent method. This demonstrates the efficiency of our proposed optimization method.

	\section{Conclusion}
	
	In this paper, we study compound weakly supervised learning motivated by real application, i.e., user experience enhancement in Didi, one of the largest online ride-sharing platforms. This is an intersection of two kinds of weak supervision, i.e., severe noisy label and distribution biased label happen simultaneously. This is a new kind of weakly supervised learning problem that to the best of our knowledge, has not been thoroughly studied before. To address this problem, we propose the \algo\ method. An instance reweighting strategy is employed to cope with severe label noise. Robust criteria  AUC and the validated performance are optimized for the correction of biased data label. Efficient algorithms accelerate the optimization on large-scale data. Experiments on real data sets validate the effectiveness of \algo\ in handling with compound weakly supervised data. In the future, we will consider extending this work to dynamic scenarios with an automatic model update.
	    
	
	\bibliography{ref.bib}
	\bibliographystyle{aaai}
\end{document}